\documentclass{isprs}
\usepackage{enumitem}
\usepackage{hyperref}
\usepackage{setspace}
\usepackage{geometry} 
\usepackage{epstopdf}
\usepackage[labelsep=period]{caption}  
\usepackage{todonotes}
\usepackage{graphicx}
\graphicspath{{images/}}
\usepackage{booktabs}
\usepackage{amsmath}
\usepackage{subcaption}
\usepackage{placeins}
\graphicspath{{images/}}

\usepackage{tummath}

\newcommand*\justify{%
  \fontdimen2\font=0.4em
  \fontdimen3\font=0.2em
  \fontdimen4\font=0.1em
  \fontdimen7\font=0.1em
  \hyphenchar\font=`\-
}

\begin{document}
\title{The SEN1-2 Dataset for Deep Learning in SAR-Optical Data Fusion}

 \author{
 M. Schmitt\textsuperscript{1}, L. H. Hughes\textsuperscript{1}, X. X. Zhu\textsuperscript{1,2} }

\address{
   \textsuperscript{1 }Signal Processing in Earth Observation, Technical University of Munich (TUM), Munich, Germany - (m.schmitt,lloyd.hughes)@tum.de\\
   \textsuperscript{2 }Remote Sensing Technology Institute (IMF), German Aerospace Center (DLR), Oberpfaffenhofen, Germany - xiao.zhu@dlr.de
}


\commission{I, }{I} 
\workinggroup{I/3} 
\icwg{}   

\abstract{\emph{This is a pre-print of a paper accepted for publication in the ISPRS Annals of the Photogrammetry, Remote Sensing and Spatial Information Sciences. Please refer to the original (open access) publication from October 2018.}\newline While deep learning techniques have an increasing impact on many technical fields, gathering sufficient amounts of training data is a challenging problem in remote sensing. In particular, this holds for applications involving data from multiple sensors with heterogeneous characteristics. One example for that is the fusion of synthetic aperture radar (SAR) data and optical imagery. With this paper, we publish the \emph{SEN1-2} dataset to foster deep learning research in SAR-optical data fusion. \emph{SEN1-2} comprises $282{,}384$ pairs of corresponding image patches, collected from across the globe and throughout all meteorological seasons. Besides a detailed description of the dataset, we show exemplary results for several possible applications, such as SAR image colorization, SAR-optical image matching, and creation of artificial optical images from SAR input data. Since \emph{SEN1-2} is the first large open dataset of this kind, we believe it will support further developments in the field of deep learning for remote sensing as well as multi-sensor data fusion.
}

\keywords{Synthetic aperture radar (SAR), optical remote sensing, Sentinel-1, Sentinel-2, deep learning, data fusion}

\maketitle

\section{Introduction}\label{sec:Intro}
Deep learning has had an enormous impact on the field of remote sensing in the past few years \cite{Zhang2016,Zhu2017}. This is mainly due to the fact that deep neural networks can model highly non-linear relationships between remote sensing observations and the eventually desired geographical parameters, which could not be represented by physically-interpretable models before. One of the most promising directions of deep learning in remote sensing certainly is its pairing with data fusion \cite{Schmitt2016}, which holds especially for a combined exploitation of \emph{synthetic aperture radar (SAR)} and optical data as these data modalities are completely different from each other both in terms of geometric and radiometric appearance. While SAR images are based on range measurements and observe physical properties of the target scene, optical images are based on angular measurements and collect information about the chemical characteristics of the observed environment. 

In order to foster the development of deep learning approaches for SAR-optical data fusion, it is of utmost importance to have access to big datasets of perfectly aligned images or image patches. However, gathering such a big amount of aligned multi-sensor image data is a non-trivial task that requires quite some engineering efforts. Furthermore, remote sensing imagery is generally rather expensive in contrast to conventional photographs used in typical computer vision applications. These high costs are mainly caused by the financial efforts associated to putting remote sensing satellite missions into space. This changed dramatically in 2014, when the SAR satellite Sentinel-1A, the first of the Sentinel missions, was launched into orbit by the European Space Administration (ESA) in the frame of the Copernicus program, which is aimed at providing an on-going supply of diverse Earth observation satellite data to the end user free-of-charge \cite{ESA2015}. 

Exploiting this novel availability of \emph{big remote sensing data}, we publish the so-called \emph{SEN1-2} dataset with this paper. It is comprised of $282{,}384$ SAR-optical patch-pairs acquired by Sentinel-1 and Sentinel-2. The patches are collected from locations spread across the land masses of the Earth and over all four seasons. The generation of the dataset, its characteristics and features, as well as some pilot applications are described in this paper.  

\section{Sentinel-1/2 Remote Sensing Data}\label{sec:background}
The Sentinel satellites are part of the Copernicus space program of ESA, which aims to replace past remote sensing missions in order to ensure data continuity for applications in the areas of atmosphere, ocean and land monitoring. For this purpose, six different satellite missions focusing on different Earth observation aspects are put into operation. Among those missions, we focus on Sentinel-1 and Sentinel-2, as they provide the most conventional remote sensing imagery acquired by SAR and optical sensors, respectively.

\subsection{Sentinel-1}\label{sec:Sen1}
The Sentinel-1 mission \cite{Torres2012} consists of two polar-orbiting satellites, equipped with C-band SAR sensors, which enables them to acquire imagery regardless of the weather.

Sentinel-1 works in a pre-programmed operation mode to avoid conflicts and to produce a consistent long-term data archive built for applications based on long time series. Depending on which of its four exclusive SAR imaging modes is used, resolutions down to 5~m with a wide coverage of up to 400~km can be achieved. Furthermore, Sentinel-1 provides dual polarization capabilities and very short revisit times of about 1 week at the equator. Since highly precise spacecraft positions and attitudes are combined with the high accuracy of the range-based SAR imaging principle, Sentinel-1 images come with high out-of-the-box geolocation accuracy \cite{Schubert2015}.   

For the Sentinel-1 images in our dataset, so-called ground-range-detected (GRD) products acquired in the most frequently available interferometric wide swath (IW) mode were used. These images contain the $\sigma^0$ backscatter coefficient in dB scale for every pixel at a pixel spacing of 5~m in azimuth and 20~m in range. For sake of simplicity, we restricted ourselves to vertically polarized (VV) data, ignoring potentially available other polarizations. Finally, for precise ortho-rectification, restituted orbit information was combined with the 30 m-SRTM-DEM or the ASTER DEM for high latitude regions where SRTM is not available.

Since we want to leave any further pre-processing to the end user so that it can be adapted to fit the desired task, we have not carried out any speckle filtering.

\subsection{Sentinel-2}\label{sec:Sen2}
The Sentinel-2 mission \cite{Drusch2012} comprises twin polar-orbiting satellites in the same orbit, phased at 180$^\circ$ to each other. The mission is meant to provide continuity for multi-spectral image data of the SPOT and LANDSAT kind, which have provided information about the land surfaces of our Earth for many decades. With its wide swath width of up to 290~km and its high revisit time of 10 days at the equator (with one satellite), and 5 days (with 2 satellites), respectively, under cloud-free conditions it is specifically well-suited to vegetation monitoring within the growing season. 

For the Sentinel-2 part of our dataset, we have only used the red, green, and blue channels (i.e. bands 4, 3, and 2) in order to generate realistically looking RGB images. Since Sentinel-2 data are not provided in the form of satellite images, but as precisely georeferenced \emph{granules}, no further processing was required. 
Instead, the data had to be selected based on the amount of cloud coverage. For the initial selection, a database query for granules with less than or equal to 1\% of cloud coverage was used.

\begin{figure*}[h]
\centering
\includegraphics[width=0.8\linewidth]{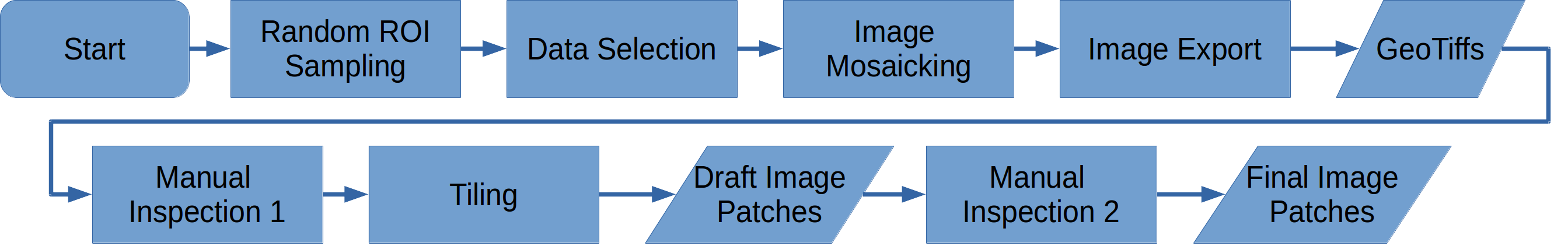}
\caption{Flowchart of the semi-automatic, Google Earth Engine-based patch extraction procedure.}\label{fig:Flowchart}
\end{figure*}

\section{The Dataset}
In order to generate a multi-sensor SAR-optical patch-pair dataset, a relatively large amount of remote sensing data with very good spatial alignment needs to be acquired. In order to do this in a mostly automatic manner, we have utilized the cloud-based remote sensing platform Google Earth Engine \cite{Gorelick2017}. The individual steps of the dataset generation procedure are described in the following.

\subsection{Data Preparation in Google Earth Engine}
The major strengths of Google Earth Engine are two-fold from the point of view of our dataset generation endeavour: On the one hand, it provides an extensive data catalogue containing several petabytes of remote sensing imagery -- including all available Sentinel data -- and other freely available geodata. On the other hand, it provides a powerful programming interface that allows to carry out data preparation and analysis tasks on Google's computing centers. Thus, we have used it to select, prepare and download the Sentinel-1 and Sentinel-2 imagery from which we have later extracted our patch-pairs. The workflow of the GEE-based image download and patch preparation is sketched in Fig.~\ref{fig:Flowchart}. In detail, it comprises the following steps:

\subsubsection{Random ROI Sampling}
In order to generate a dataset that represents the versatility of our Earth as good as possible, we wanted to sample the scenes used as basis for dataset production over the whole globe. For this task, we use Google Earth Engine's \texttt{\justify ee.FeatureCollection.randomPoints()} function to randomly sample points from a uniform spatial distribution. Since many remote sensing investigations focus on urban areas and since urban areas contain more complex visual patterns than rural areas, we introduce a certain artificial bias to urban areas by sampling 100 points over all land masses of the Earth and another 50 points only over urban areas. The shape files for both land masses and urban areas were provided by the public domain geodata service \url{www.naturalearthdata.com} at a scale of 1:50m. If two points are located in close proximity to each other, we removed one of them to ensure non-overlapping scenes. 

\begin{figure*}[ht]
\centering
\begin{subfigure}[c]{0.45\linewidth}
\centering
\includegraphics[width=0.9\linewidth]{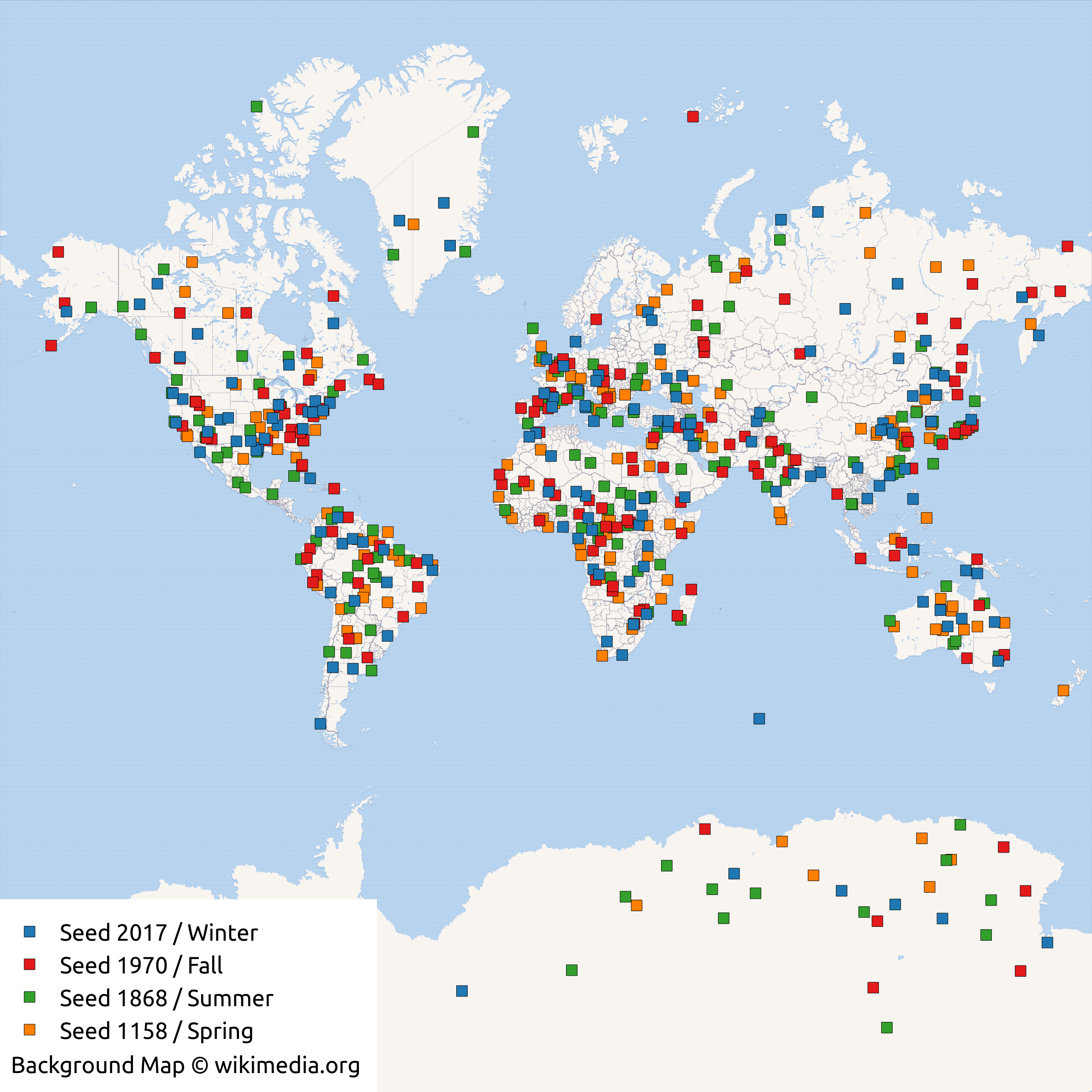}
\subcaption{}\label{fig:WorldMap_full}
\end{subfigure}
\begin{subfigure}[c]{0.45\linewidth}
\centering
\includegraphics[width=0.9\linewidth]{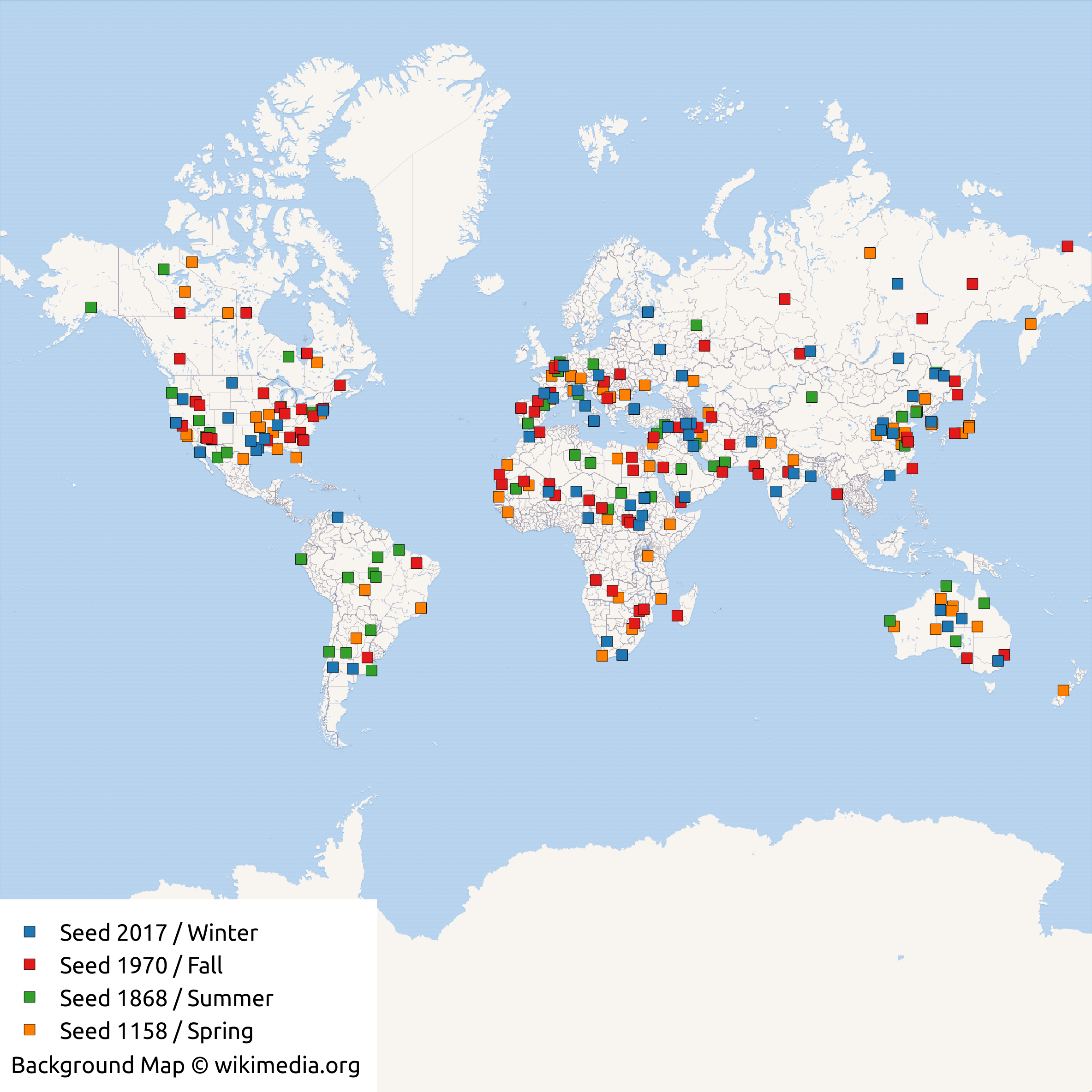}
\subcaption{}\label{fig:WorldMap_reduced}
\end{subfigure}
\caption{Distribution of the ROIs sampled uniformly over the land masses of the Earth: (a) Original ROIs, (b) final set of scenes after removal of cloud- and/or artifact-affected ROIs.}\label{fig:WorldMap}
\end{figure*}

This sampling process is carried out for four different seed values (1158, 1868, 1970, 2017). The result of the random ROI sampling is illustrated in Fig.~\ref{fig:WorldMap_full}.

\subsubsection{Data Selection}
\begin{figure*}[th]
\centering
\includegraphics[width=0.65\linewidth]{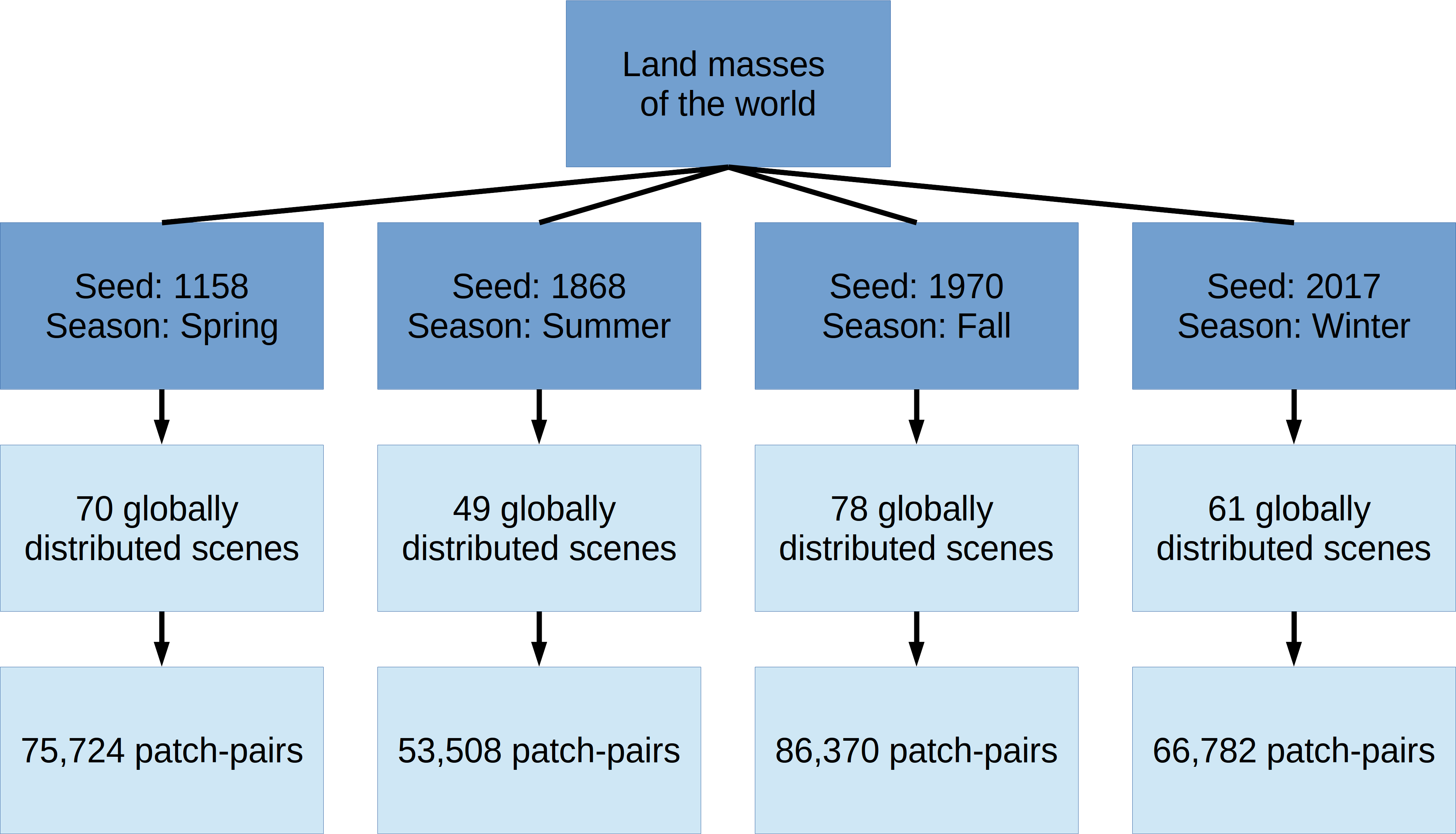}
\caption{Structure of the final dataset.}\label{fig:DatasetStructure}
\end{figure*}

In the second step, we use GEE's tools to filter image collections to select the Sentinel-1/Sentinel-2 image data for our scenes. Since we want to use only recent data acquired in 2017, this first means that we structure the year into the four meteorological seasons: winter (1 December 2016 to 28 February 2017), spring (1 March 2017 to 30 May 2017), summer (1 June 2017 to 31 August 2017), and fall (1 September 2017 to 30 November 2017). Each season is then associated to one of the four sets of random ROIs, thus providing us with the top-level dataset structure (cf. Fig.~\ref{fig:DatasetStructure}): We structure the final dataset into four distinct sub-groups ROIs1158\_spring, ROIs1868\_summer,\newline ROIs1970\_fall, and ROIs2017\_winter. 

Then, for each ROI, we filter for Sentinel-2 images with a maximum cloud coverage of 1\% and for Sentinel-1 images acquired in IW mode with VV polarization. If no cloud-free Sentinel-2 image or no VV-IW Sentinel-1 image is available within the corresponding season, the ROI is discarded. Thus, the number of ROIs is significantly reduced from about $600$ to about $429$. For example, all ROIs that were located in Antarctica are rendered obsolete, since the geographical coverage of Sentinel-2 is restricted to $56^\circ$ South to $83^\circ$ North.

\subsubsection{Image Mosaicking}
Continuing with the selected image data, we use the Google Earth Engine in-built functions \texttt{ee.Image-Collection.mosaic()} and \texttt{ee.Image.clip()} to create one single image for each ROI, clipped to the respective ROI extent. The \texttt{ee.ImageCollection.mosaic()} function simply composites overlapping images according to their order in the collection in a \emph{last-on-top} sense. As mentioned in Section~\ref{sec:Sen2}, we select only bands 4, 3, and 2 for Sentinel-2 in order to create RGB images. 

\subsubsection{Image Export}
Finally, we export the images created in the previous steps as GeoTiffs using the GEE function \texttt{Export.image.toDrive} and a scale of 10m. The downloaded GeoTiffs are then pre-processed for further use by cutting the gray values to the $\pm2.5\sigma$ range, scaling them to the interval $[0;1]$ and performing a contrast-stretch. These corrections are applied to all bands individually.  

\subsubsection{First Manual Inspection}
We then visually inspect all downloaded scenes for severe problems. These can mostly belong to one of the following categories:
\begin{itemize}[itemsep=1pt,topsep=-5pt]
\item Large \emph{no-data} areas.\newline 
Unfortunately, the \texttt{ee.ImageCollection.mosaic()} function does not return any error message if it does not find a suitable image to fill the whole ROI with data. This mostly happens to Sentinel-2, when no sufficiently cloud-free granule is available for a given time period.
\item Strong cloud coverage.\newline
The cloud-coverage metadata information that comes with every Sentinel-2 granule is only a global parameter. Thus, it can happen that the whole granule only contains a few clouds, but the part covering our ROI is where all the clouds reside. 
\item Severely distorted colors.\newline
Sometimes, we observed very unnatural colors for Sentinel-2 images. Since we want to create a dataset that contains naturally looking RGB images for Sentinel-2, we also removed some Sentinel-2 images with all too strange colors. 
\end{itemize}
After this first manual inspection, only $258$ scenes/ROIs remain (cf. Fig.~\ref{fig:WorldMap_reduced}). 

\subsubsection{Tiling}
Since our goal is a dataset of patch-pairs that can be used to train machine learning models aiming at various data fusion tasks, we eventually seek to generate patches of $256 \times 256$ pixels. Using a stride of 128, we reduce the overlap between neighboring patches to only $50\%$ while maximising the number of independent patches we can get out of the available scenes. We end up with $298{,}790$ Sentinel-1/Sentinel-2 patch-pairs after this step.

\begin{figure*}[!h]
\centering
\setlength\tabcolsep{1.5pt}
\begin{tabular}{ccccccccc}
\includegraphics[width=0.12\textwidth]{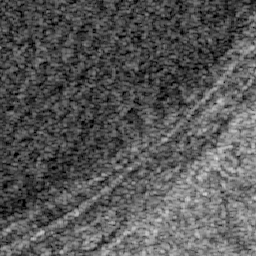} & \includegraphics[width=0.12\textwidth]{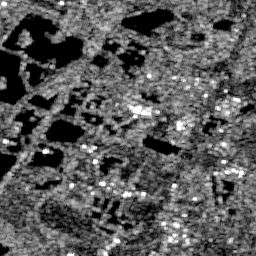} &
\includegraphics[width=0.12\textwidth]{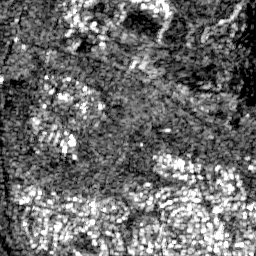} &
\includegraphics[width=0.12\textwidth]{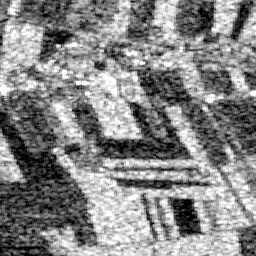} &
\includegraphics[width=0.12\textwidth]{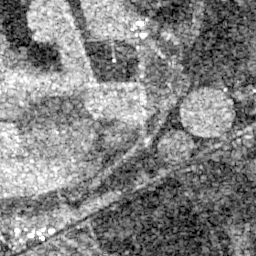} &
\includegraphics[width=0.12\textwidth]{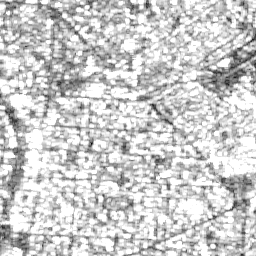} &
\includegraphics[width=0.12\textwidth]{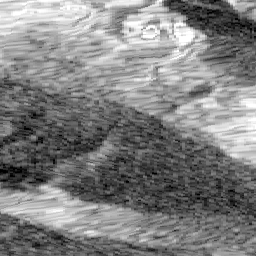} &
\includegraphics[width=0.12\textwidth]{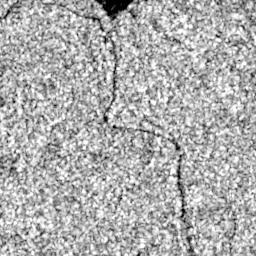}\\
\includegraphics[width=0.12\textwidth]{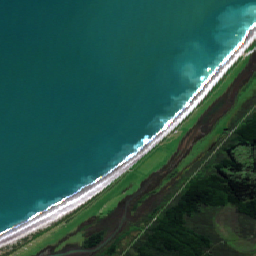} & \includegraphics[width=0.12\textwidth]{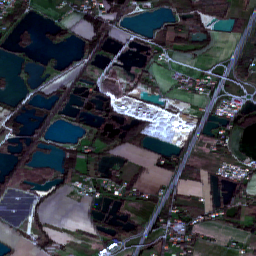} &
\includegraphics[width=0.12\textwidth]{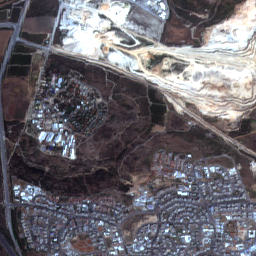} &
\includegraphics[width=0.12\textwidth]{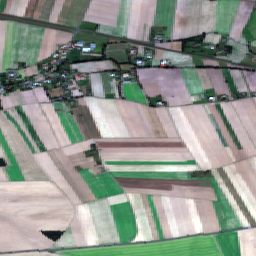} &
\includegraphics[width=0.12\textwidth]{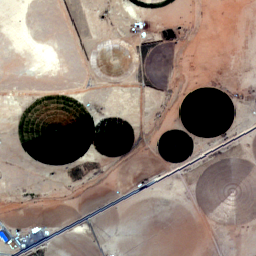} &
\includegraphics[width=0.12\textwidth]{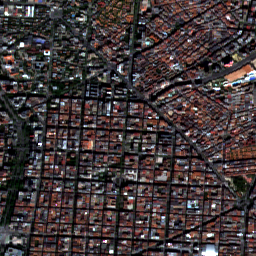} &
\includegraphics[width=0.12\textwidth]{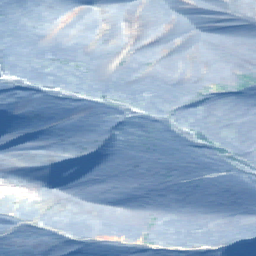} &
\includegraphics[width=0.12\textwidth]{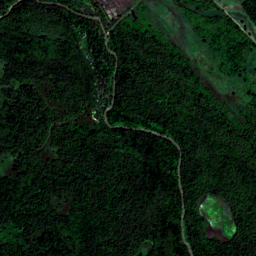}
\end{tabular}
\caption{Some exemplary patch-pairs from the \emph{SEN1-2} dataset. Top row: Sentinel-1 SAR image patches, bottom row: Sentinel-2 RGB image patches.}\label{fig:examples}
\end{figure*}

\subsubsection{Second Manual Inspection}
In order to remove sub-\\optimal patches that, e.g., still contain small clouds or visible mosaicking seamlines, we have again inspected all patches visually. In this step, $16{,}406$ patch-pairs are manually removed, leaving the final amount of $282{,}384$ quality-controlled patch-pairs. Some examples are shown in Fig.~\ref{fig:examples}.

\subsection{Dataset Availability}
The \emph{SEN1-2} dataset is shared under the open access license CC-BY and available for download at a persistent link provided by the library of the Technical University of Munich: \url{https://mediatum.ub.tum.de/1436631}. This paper must be cited when the dataset is used for research purposes. 

\section{Example Applications}
In this section, we present some example applications, for which the dataset has been used already. These should serve as inspiration for future use cases and ignite further research on SAR-optical deep learning-based data fusion.

\subsection{Colorizing Sentinel-1 Images}
\begin{figure}[!h]
\centering
\setlength\tabcolsep{1.5pt}
\begin{tabular}{cccc}
\includegraphics[width=0.115\textwidth]{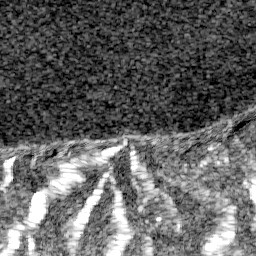} & 
\includegraphics[width=0.115\textwidth]{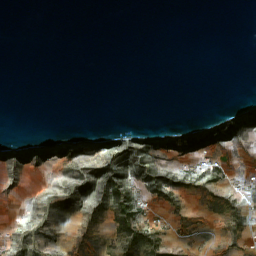} &
\includegraphics[width=0.115\textwidth]{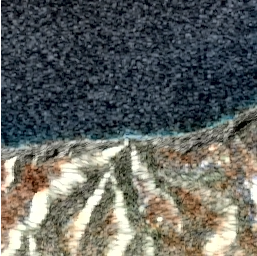} &
\includegraphics[width=0.115\textwidth]{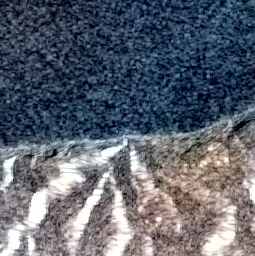}\\
\includegraphics[width=0.115\textwidth]{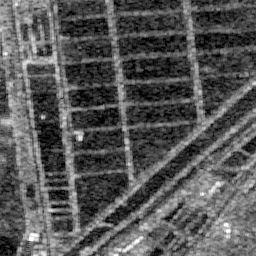} &
\includegraphics[width=0.115\textwidth]{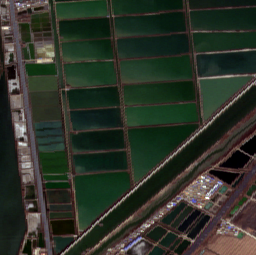} &
\includegraphics[width=0.115\textwidth]{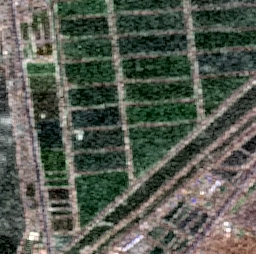} &
\includegraphics[width=0.115\textwidth]{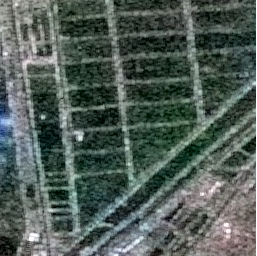}\\
\includegraphics[width=0.115\textwidth]{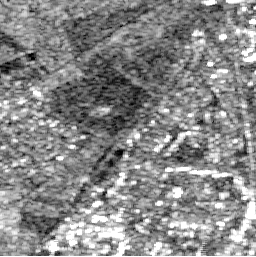} &
\includegraphics[width=0.115\textwidth]{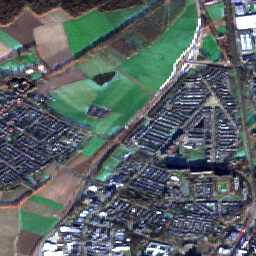} &
\includegraphics[width=0.115\textwidth]{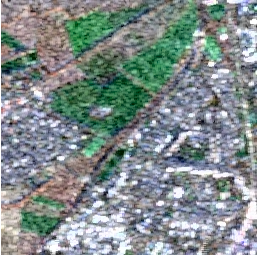} &
\includegraphics[width=0.115\textwidth]{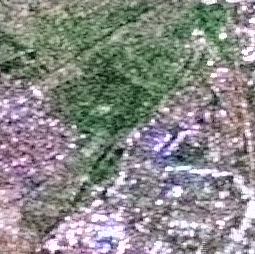}
\end{tabular}
\caption{Some results for colorized SAR image patches. In each row, from left to right: original Sentinel-1 SAR image patch, corresponding Sentinel-2 RGB image patch, artificial color SAR patch based on color-space-based SAR-optical image fusion, artificial color SAR image predicted by a deep generative model.}\label{fig:colorExamples}
\end{figure}

The interpretation of SAR images is still a highly non-trivial task, even for well-trained experts. One reason for this is the missing color information, which supports any human image understanding endeavour. One promising field of application for the \emph{SEN1-2} dataset thus is to learn to colorize gray-scale SAR images with color information derived from corresponding optical images, as we have proposed earlier \cite{Schmitt2018}. 
In this approach, we make use of SAR-optical image fusion to create artificial color SAR images as training examples, and of the combination of variational autoencoder and mixture density network proposed by \cite{Deshpande17} to learn a conditional color distribution, from which different colorization samples can be drawn. Some first results resulting from a training on $252{,}384$ \emph{SEN1-2} patch pairs are displayed in Fig.~\ref{fig:colorExamples}. 

\subsection{SAR-optical Image Matching}
\begin{figure}[!h]
\centering
\setlength\tabcolsep{1.5pt}
\begin{tabular}{cccc}
\includegraphics[width=0.115\textwidth]{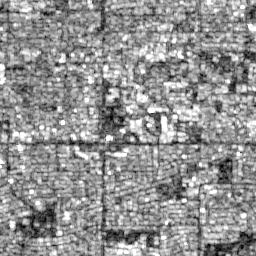} & 
\includegraphics[width=0.115\textwidth]{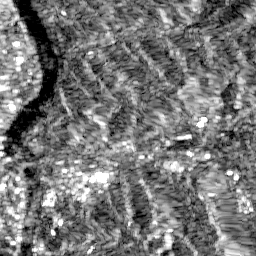} &
\includegraphics[width=0.115\textwidth]{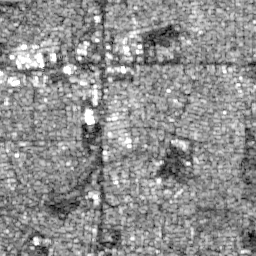} &
\includegraphics[width=0.115\textwidth]{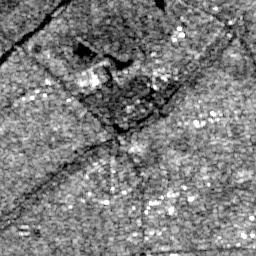}\\
\includegraphics[width=0.115\textwidth]{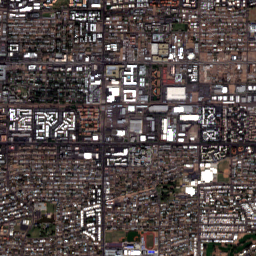} &
\includegraphics[width=0.115\textwidth]{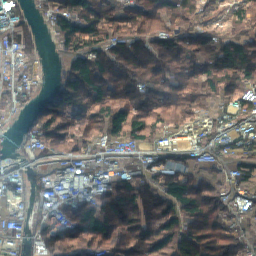} &
\includegraphics[width=0.115\textwidth]{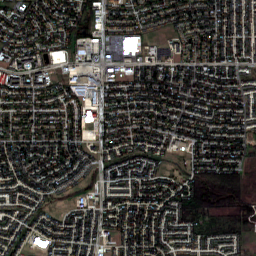} &
\includegraphics[width=0.115\textwidth]{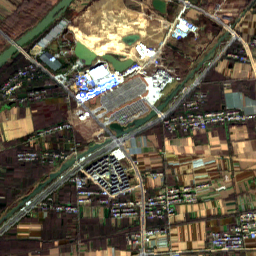}
\end{tabular}

\caption{Some true positives achieved in SAR-optical image matching. The first row depicts the Sentinel-1 SAR image patch, while the second row depicts the corresponding Sentinel-2 optical patch as predicted by a pseudo-siamese convolutional neural network.}\label{fig:matchingExamples}
\end{figure}

Tasks such as image co-registration, 3D stereo reconstruction, or change detection rely on being able to accurately determine similarity (i.e. matching) between corresponding parts in different images. While well-established methods and similarity measures exist to achieve this for mono-modal imagery, the matching of multi-modal data remains challenging to this day. The \emph{SEN1-2} dataset can assist in creating solutions in the field of multi-modal image matching by providing the large quantities of data required to exploit modern \emph{deep matching} approaches, such as proposed by \cite{Merkle2017} or \cite{Hughes2018}: Using a pseudo-siamese convolutional neural network architecture, corresponding SAR-optical image patches of a \emph{SEN1-2} test subset can be identified with an accuracy of $93\%$. The confusion matrix for the model of \cite{Hughes2018} trained on $300{,}000$ corresponding and non-corresponding patch pairs created from a \emph{SEN1-2} training subset can be seen in Tab.~\ref{tab:matchingExamples}. Furthermore, some exemplary matches achieved on the test subset are shown in Fig.~\ref{fig:matchingExamples}. 

\begin{table}[h!]
\centering
\caption{Confusion Matrix for Pseudo-siamese patch matching trained on \emph{SEN1-2}}
\label{tab:matchingExamples}
\begin{tabular}{|l|l|l|}
\hline
$\hat{y}/y$ & non-match & match   \\ \hline
non-match   & 93.84\%   & 6.16\%  \\ \hline
match       & 6.02\%    & 93.98\% \\ \hline
\end{tabular}
\end{table}

\subsection{Generating Artificial Optical Images from SAR Inputs}
\begin{figure}[!h]
\centering
\setlength\tabcolsep{3pt}
\begin{tabular}{ccc}
\includegraphics[width=0.11\textwidth]{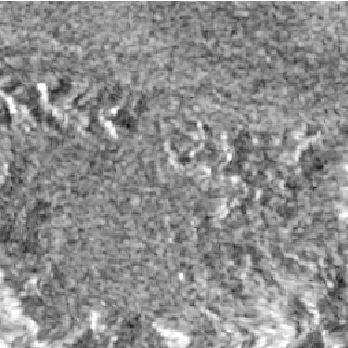} & 
\includegraphics[width=0.11\textwidth]{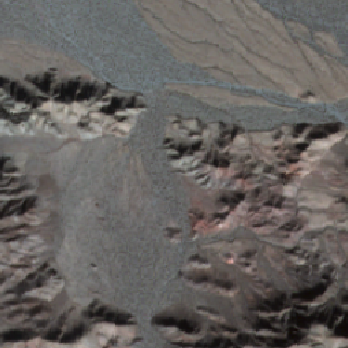} &
\includegraphics[width=0.11\textwidth]{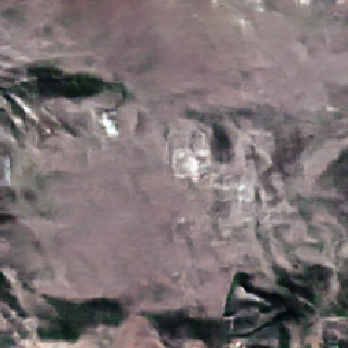}\\
\includegraphics[width=0.11\textwidth]{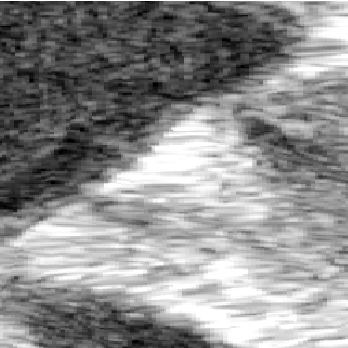} &
\includegraphics[width=0.11\textwidth]{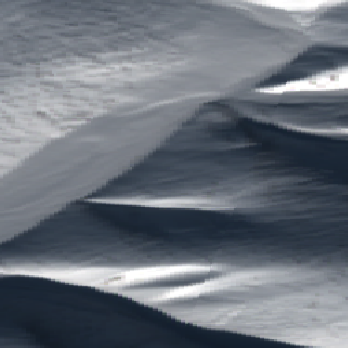} &
\includegraphics[width=0.11\textwidth]{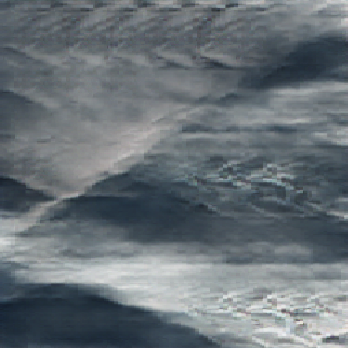}\\
\includegraphics[width=0.11\textwidth]{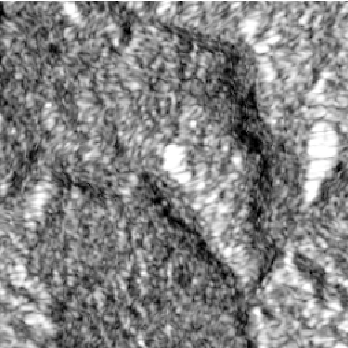} &
\includegraphics[width=0.11\textwidth]{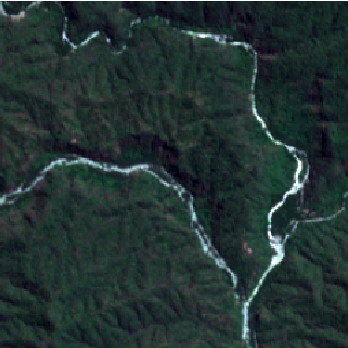} &
\includegraphics[width=0.11\textwidth]{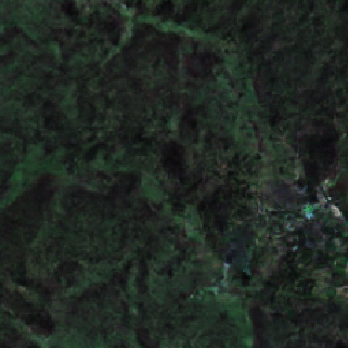}
\end{tabular}
\caption{Some preliminary examples for the prediction of artificial optical images from SAR input data using the \texttt{pix2pix} GAN. In each row, from left to right: original Sentinel-1 SAR image patch, corresponding Sentinel-2 RGB image patch, artificial GAN-predicted optical image patch.}\label{fig:pix2pix}
\end{figure}

Another possible field of application of the \emph{SEN1-2} dataset is to train generative models that allow to predict artificial SAR images from optical input data \cite{Marmanis2017,Merkle2018} or artificial optical imagery from SAR inputs \cite{Wang2018b,Ley2018,Grohnfeldt2018}. Some preliminary examples based on the well-known generative adversarial network (GAN) \texttt{pix2pix} \cite{Isola2017} trained on $108{,}221$ \emph{SEN1-2} patch pairs are shown in Fig. \ref{fig:pix2pix}.

\section{Strengths and Limitations of the Dataset}
To our knowledge, \emph{SEN1-2} is the first dataset providing a really large amount ($>100{,}000$) of co-registered SAR and optical image patches. The only other existing dataset in this domain is the so-called SARptical dataset published by \cite{Wang2018}. In contrast to the \emph{SEN1-2} dataset, it provides very-high-resolution image patches from TerraSAR-X and aerial photogrammetry, but is restricted to a mere $10{,}000$ patches extracted from a single scene, which is possibly not sufficient for many deep learning applications -- especially since many patches show an overlap of more than $50\%$. With its $282{,}384$ patch-pairs spread over the whole globe and all meteorological seasons, \emph{SEN1-2} will thus be a valuable data source for many researchers in the field of SAR-optical data fusion and remote sensing-oriented machine learning. A particular advantage is that the dataset can easily be split into various deterministic subsets (e.g. according to scene or according to season), so that truly independent training and testing datasets can be created, supporting unbiased evaluations with regard to unseen data. 

However, also \emph{SEN1-2} does not come without limitations: For example, we restricted ourselves to RGB images for the Sentinel-2 data, which is possibly insufficient for researchers working on the exploitation of the full radiometric bandwidth of multi-spectral satellite imagery. Furthermore, at the time we carried out the dataset preparation, GEE stocked only Level-1C data for Sentinel-2, which basically means that the pixel values represent top-of-atmosphere (TOA) reflectances instead of atmospherically corrected bottom-of-atmosphere (BOA) information. We are planning to extend the dataset for a future version 2 release accordingly.

\section{Summary and Conclusion}
With this paper, we have described and released the \emph{SEN1-2} dataset, which contains $282{,}384$ pairs of SAR and optical image patches extracted from versatile Sentinel-1 and Sentinel-2 scenes. We assume this dataset will foster the development of machine learning, and in particular, deep learning approaches in the field of satellite remote sensing and SAR-optical data fusion. For the future, we plan on releasing a refined, second version of the dataset, which contains not only RGB Sentinel-2 images, but full multi-spectral Sentinel-2 images including atmospheric correction. In addition, we might add coarse land use/land cover (LULC) class information to each patch-pair in order to foster also developments in the field of LULC classification.

\section*{Acknowledgements}
This work is jointly supported by the Helmholtz Association under the framework of the Young Investigators Group SiPEO (VH-NG-1018, www.sipeo.bgu.tum.de), the German Research Foundation (DFG) as grant SCHM 3322/1-1, and the European Research Council (ERC) under the European Union's Horizon 2020 research and innovation programme (grant agreement ERC-2016-StG-714087, Acronym: \textit{So2Sat}).

{
	\begin{spacing}{0.9}
		\bibliography{bib/references} 
	\end{spacing}
}


\end{document}